\newcommand{\ignore}[1]{}

\documentclass[10pt, conference]{IEEEtran}

\usepackage{pdfpages}
\usepackage{listings}
\usepackage{url}
\usepackage{multirow}
\usepackage{subfigure}
\usepackage{epsfig}
\usepackage{graphicx}
\usepackage{amsmath}
\usepackage{amssymb}
\usepackage{color}
\usepackage{setspace}
\usepackage{cite}
\usepackage{hhline}   
\usepackage{algorithm}
\usepackage{algpseudocode}
\usepackage{flushend}
\usepackage{url}
\usepackage[normalem]{ulem}
\usepackage{graphics}
\usepackage{tikz}
\usepackage{array}% http://ctan.org/pkg/array
\usepackage{animate}
\usepackage{titlesec}
\usepackage{balance}
\usepackage{enumitem}
\usepackage{booktabs}

\titlespacing*{\section}{0pt}{3pt}{3pt}
\titlespacing*{\subsection}{0pt}{3pt}{2pt}
\titlespacing*{\subsubsection}{0pt}{3pt}{2pt}

\definecolor{codegreen}{rgb}{0,0.6,0}
\definecolor{codegray}{rgb}{0.5,0.5,0.5}
\definecolor{codepurple}{rgb}{0.58,0,0.82}
\definecolor{backcolour}{rgb}{0.95,0.95,0.92}
\definecolor{Burgundy1}{RGB}{159,29,53}
\lstdefinestyle{mystyle}{
    commentstyle=\color{codegreen}\itshape,
    keywordstyle=\color{black},
    numberstyle=\tiny\color{codegray},
    stringstyle=\color{Burgundy1},
    basicstyle=\scriptsize,
    breakatwhitespace=false,         
    breaklines=true,                 
    captionpos=b,                    
    keepspaces=true,                 
    backgroundcolor=\color{backcolour},   
    numbersep=2pt,                  
    showspaces=false,                
    showstringspaces=false,
    showtabs=false,                  
    tabsize=2
}

\def\dataset{EDA Corpus}

\begin{document}
% I have listed two options here. The second one is slightly better I think. Austin, feel free to let me know what you think.
% \title{\dataset{}: A Prompt and Response \\ Dataset for OpenROAD}
\title{\dataset{}: A Large Language Model Dataset for Enhanced Interaction with OpenROAD}
%\title{OpenROAD Assistant Dataset: A Large Language Model Dataset for Enhanced Interaction with OpenROAD}

%\title{}
\author{Bing-Yue Wu$^1$, Utsav Sharma$^2$, Sai Rahul Dhanvi Kankipati$^1$, Ajay Yadav$^1$, Bintu Kappil George$^1$, \\ Sai Ritish Guntupalli$^1$, Austin Rovinski$^2$, and Vidya A. Chhabria$^1$ \\ $^1$Arizona State University; $^2$New York University}

\maketitle

% 200 word limit.
\begin{abstract}
\noindent
Large language models (LLMs) serve as powerful tools for design, providing capabilities for both task automation and design assistance. Recent advancements have shown tremendous potential for facilitating LLM integration into the chip design process; however, many of these works rely on data which are not publicly available and/or not permissively licensed for use in LLM training and distribution.
In this paper, we present a solution aimed at bridging this gap by introducing an open-source dataset tailored for OpenROAD, a widely adopted open-source EDA toolchain. The dataset features over 1000 data points and is structured in two formats: (i) a pairwise set comprised of question prompts with prose answers, and (ii) a pairwise set comprised of code prompts and their corresponding OpenROAD scripts. By providing this dataset, we aim to facilitate LLM-focused research within the EDA domain. The dataset is available at https://github.com/OpenROAD-Assistant/EDA-Corpus.

% Old
%Recent developments have shown the potential of large language models (LLMs) as assistants for utilizing Electronic Design Automation (EDA) tools and facilitating chip design. Despite these advancements, the integration of LLMs into EDA faces notable obstacles, primarily due to the absence of an open infrastructure that encompasses open datasets, benchmarks, and unrestricted open-source EDA tools. This paper presents a solution aimed at bridging this gap by introducing an open-source dataset tailored for OpenROAD, a widely adopted open-source EDA toolchain. The dataset, featuring over 1000 data points, is structured in two formats: (i) a set comprising prompts and their corresponding OpenROAD-generated scripts, and (ii) a collection containing questions as prompts with prose answers. By providing this dataset, we aim to facilitate LLM-focused research within the EDA domain.
\end{abstract}

\section{Introduction} 
\label{sec:intro}
\noindent
Chip design is a complex process which requires deep domain expertise, both in foundational knowledge and in the electronic design automation (EDA) tools used in creating chips. This expertise creates a barrier not only for newcomers to chip design, but even for experts switching to different tools or sub-domains within chip design, lowering their productivity.
Recent advancements in LLMs have demonstrated tremendous potential in task automation and comprehension of esoteric topics. Works like ChatGPT~\cite{ChatGPT}, Copilot~\cite{copilot}, and others have shown high performance on a wide variety of tasks; however, these LLMs are powered by training on extremely large data corpora. General-purpose LLMs are known to experience trouble extending to esoteric tasks when their training corpora lack sufficient coverage of the target domain.

Recent works on domain-specific LLMs for EDA have shown that tailored data sets and fine-tuning mechanisms can significantly improve performance over foundation models~\cite{ChatEDA, LLM4EDA, liu2024chipnemo}. While models which relate to frontend RTL design have significant data available due to open-source languages such as Verilog, models related to backend physical design have scarcely any data available due to a heavy reliance on proprietary APIs, tools, and data.

To date, LLM methodologies in physical design EDA face significant challenges due to the absence of accessible, open infrastructure. The high cost of licenses for access to commercial EDA tools, documentation, and tutorials places them out of reach for the wider community. Moreover, these tools are bound by strict end-user license agreements which often restrict activities such as benchmarking, training AI with their documentation, and freely exchanging user scripts.

To foster research in LLM-assisted physical design, we introduce \dataset{}, a curated dataset for physical design automation tasks. \dataset{} is based on OpenROAD~\cite{ajayi2019openroad}, a widely utilized open-source EDA tool for automated place and route tasks. Leveraging OpenROAD mitigates obstacles associated with proprietary EDA tools, enabling the public release of our dataset and facilitating its use with LLMs without licensing constraints.
\dataset{} consists of two types of data: (i) question and answer pairs,  and (ii) prompt and script pairs. The question-answer dataset contains pairs of prose questions about OpenROAD and the corresponding prose answer. The prompt-script dataset contains pairs of prose requests to execute actions and the corresponding Python script which executes the actions using the OpenROAD API.

\noindent
Our key contributions include the following:
\begin{itemize}
    \item Release of a question-answer dataset to train LLMs on answering questions about physical design methods using OpenROAD. Each data point was provided by or verified as accurate by OpenROAD experts.
    \item Release of a prompt-script dataset to train LLMs for script generation for physical design tasks in OpenROAD. Each data point was verified through execution in OpenROAD.
    \item Demonstration of improvement over state-of-the-art LLMs by fine-tuning ChatGPT~\cite{ChatGPT} with \dataset{}.
\end{itemize}

To the best of our knowledge, this is the first publicly released~\cite{github}, permissively licensed dataset to drive training of LLMs for physical design tasks.
By releasing this dataset, we aim to seed efforts for training LLM assistants for physical design. LLMs have the capability to dramatically increase physical design accessibility for both new and seasoned chip designers alike.
Furthermore, it sets an exemplar for continued research in EDA and physical design.

\section{Background}
\label{sec:preliminaries}

\subsection{LLMs for Chip Design and EDA}
\label{sec:dataset_for_llm_in_IC}
Recently, there have been several works on using LLMs for hardware design that span high-level synthesis~\cite{10473893HLS}, hardware description language (HDL) generation, place and route script generation~\cite{ChatEDA}, and more. LLM4EDA~\cite{LLM4EDA} establishes a taxonomy for three classes of EDA LLMs:

\begin{itemize}
    \item \textbf{Chatbot assistants} for learning the details of an EDA tool flow and process to design chips~\cite{liu2024chipnemo}
    \item \textbf{Code generation and evaluation} for generating high-quality HDL code~\cite{thorat2024advanced,chang2023chipgpt,thakur2023autochip,10323812verilogeval,fu2023gpt4aigchip} or EDA tool scripts~\cite{ChatEDA}
    %Aiming to reduce human errors and accelerate the design process.
    \item \textbf{HDL verification and analysis} for code summarizing and bug-finding~\cite{liu2024chipnemo,tsai2024rtlfixer,orenesvera2023using,meng2023unlocking}
\end{itemize}

\noindent
To the best of our knowledge, prior art using LLMs for physical design is extremely limited. ChatEDA~\cite{ChatEDA} conducts fine-tuning using 1,500 self-generated data points; however, there is no public dataset that has been released as a part of this work. To fill this gap, \dataset{} provides datasets for training LLMs on physical design tasks within OpenROAD. 
It aims to train chatbots to assist in both understanding physical design in the context of OpenROAD as well as converting user intent into physical design actions. The corpus encompasses two of the three categories listed above: \emph{\textbf{assistant chatbot}} and \emph{\textbf{script generation}}, which are described further in Section~\ref{sec:dataset}.

\subsection{Relevance of OpenROAD for training LLMs}
\label{sec:python_apis}
Open-source physical design platforms such as OpenROAD~\cite{ajayi2019openroad} and iEDA~\cite{li2023ieda} have been widely used for education and research. OpenROAD, is an open-source platform for RTL to GDS and is composed of several different tools tightly integrated together into a single application via a database, OpenDB~\cite{OpenDB}. 
Unlike commercial tools, which work with Tcl interfaces, OpenROAD has Python interfaces~\cite{VTS} which are crucial in training LLMs for script generation.

Although OpenROAD has been widely used, it does pose challenges for first time users including installation-related issues and  understanding how the underlying EDA tools work. Even though there are Python-based user interfaces which are simpler to use compared to Tcl, understanding each API and writing scripts using these APIs to query/modify the database and perform physical design tasks is challenging due to the limited documentation available and the requirement to be familiar with OpenROAD source code. 
We aim to overcome the above challenges by providing a dataset that can be used to train LLMs to serve as chatbots and EDA tool script generators to help users of OpenROAD improve productivity.

\section{Dataset Description}
\label{sec:dataset}
\begin{figure}[t]
    \centering
    \includegraphics[width= 0.92\linewidth]{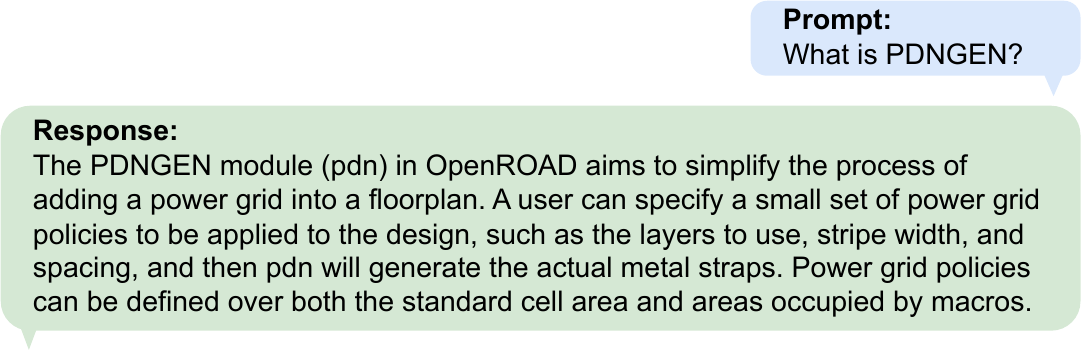}
    \vspace{-3mm}
    \caption{Sample knowledge-based question-answer datapoint.}
    \label{fig:eq_example}
\end{figure}

\begin{figure}[t]
    \centering
    \includegraphics[width= 0.92\linewidth]{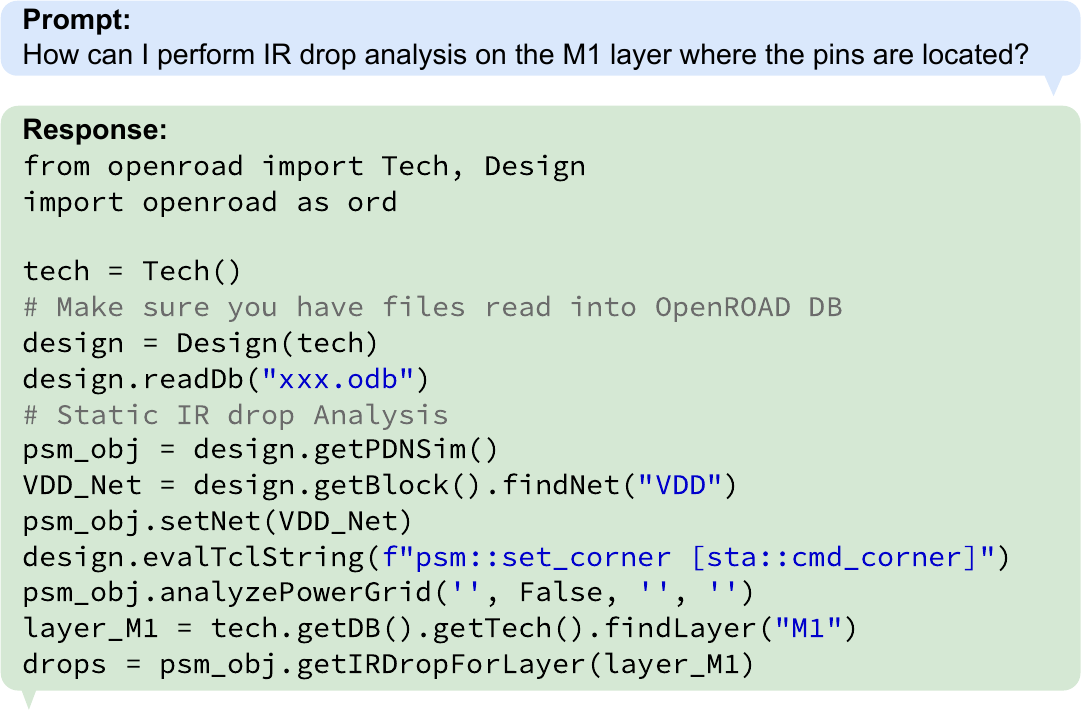}
    \vspace{-3mm}
    \caption{Sample flow-based prompt-script datapoint for IR drop analysis.}
    \vspace{-5mm}
    \label{fig:code_example}
\end{figure}

\noindent
This paper introduces a dataset which consists of prompt-response pair data, encompassing (i) a prose dataset which includes knowledge-based question-answer pairs, as shown in Fig.~\ref{fig:eq_example}, and (ii) a script dataset which includes operation-based prompt-script pairs, as shown in Fig.~\ref{fig:code_example}.

\subsection{Question-Answer Dataset}
\noindent
The question-answer dataset contains pairs of question prompts and prose answers which are collected from the OpenROAD GitHub issues, discussions, and documentation. Data from the issues and discussions were sampled from real user questions with answers provided by OpenROAD experts. Data from the documentation was created by generating questions corresponding to answers present in the OpenROAD documentation.
The datapoints are categorized into three categories: OpenROAD general, OpenROAD tool, and OpenROAD flow.

\noindent
\textbf{General:} Questions in this category relate to general capabilities of OpenROAD, such as high-level questions, common debugging questions, or design strategy questions.

\noindent
\textbf{Tool:} Questions in this category relate to specific OpenROAD tools, including algorithm questions, command argument questions, and result interpretation questions.

\noindent
\textbf{Flow:} Questions in this category relate specifically to the OpenROAD Flow, which is the user-facing RTL-to-GDS flow maintained by The OpenROAD Project.

\subsection{Prompt-Script Dataset}
The prompt-script dataset is comprised of prompts and OpenROAD Python code pairs. While Tcl is the normal interface for OpenROAD, leveraging Python allows the reuse of pretrained LLMs for Python code generation. It is worth noting that there are currently no publicly available LLMs trained specifically for generating Tcl scripts, hence the focus on Python-based scripts. Within our prompt-script dataset, data points are categorized into two main types: flow-based scripts and database (DB)-based scripts.

\noindent
\textbf{Flow-based:} This category of our script dataset includes those datapoints that perform physical design tasks, such as floorplanning, placement, routing, and essential OpenROAD-supported procedures, e.g., reading and writing for files and database representation of the design.

\noindent
\textbf{DB-based:} This category of our script dataset includes those data points that directly interact with the DB to either query the DB for information or modify the netlist and layout. These datapoints utilize the ``get" and ``set" helper functions of the DB.
Examples include extracting design information, such as cell, net, and pin properties, netlist modification APIs, such as gate sizing and buffering, and incremental changes to layout, such as flipping a cell or rotating it. We further subcategorize this into query-based datapoints (which utilize several "get" helper functions in OpenDB) and modification-based datapoints that incrementally change the database~\cite{github}.

\section{Dataset evaluation: Quality and statistics}

\begin{table}
\centering
\caption{Question-answer and prompt-script dataset statistics}
\label{tbl:stats}
\resizebox{\linewidth}{!}{%
\begin{tabular}{||l|l||r|r||r||} 
\hhline{|t:==:t:==:t:=:t|}
\multicolumn{1}{||c|}{\textbf{Dataset}} & \multicolumn{1}{c||}{\textbf{Category}} & \multicolumn{1}{c|}{\textbf{Unique}} & \multicolumn{1}{c||}{\textbf{Augmented}} & \multicolumn{1}{c||}{\textbf{Total}} \\ 
\hhline{|:==::==::=:|}
\multirow{3}{*}{\textbf{Question-answer }} & General & 61 & 120 & 181 \\ 
\cline{2-5}
 & Tools & 64 & 126 & 190 \\ 
\cline{2-5}
 & Flow & 73 & 146 & 219 \\ 
\hhline{|:==::==::=:|}
\multicolumn{2}{||l||}{\textbf{Total question-answer }} & 198 & 392 & 590 \\ 
\hhline{|:==::==::=:|}
\multirow{2}{*}{\textbf{Prompt-script }} & Flow & 138 & 235 & 373 \\ 
\cline{2-5}
 & Database & 258 & 312 & 570 \\ 
\hhline{|:==::==::=:|}
\multicolumn{2}{||l||}{\textbf{Total prompt-script }} & 395 & 545 & 943 \\
\hhline{|b:==:b:==:b:=:b|}
\end{tabular}
}
\vspace{-4mm}
\end{table}

\subsection{Curating the Dataset}
\label{sec:collecting_data}
\noindent
{\bf Question-answer dataset}
The most significant challenge in creating this dataset, and one of our major contributions, is curating this data to filter out noise. Many questions and answers in the source data (i) do not have an answer clearly marked on GitHub, (ii) were resolved through OpenROAD bugfixes (and therefore are not relevant), or (iii) contain interspersed irrelevant conversation. The question-answer pairs in this dataset were curated to provide concise questions with direct, generalized answers and remove spurious data pairs. Prior LLM works have shown that even small amounts of high-quality training data can produce high-quality models through fine-tuning foundation models~\cite{zhou2024lima}. Given the low availability of question-answer data for EDA, providing a curated, high-quality dataset is essential to facilitate future research.

Table~\ref{tbl:stats} identifies the distribution of data points based on the categories mentioned in Section~\ref{sec:dataset}. The dataset features 198 unique data points as question-answer pairs. In addition, this dataset is supplemented by augmented data pairs formed through paraphrasing questions and answers. Between the unique and augmented data pairs, the dataset comprises nearly 600 data points.

\noindent {\bf Prompt-script dataset}
Table~\ref{tbl:stats} presents the distribution of datapoints across categories in our prompt-script dataset. The table distinguishes between unique-functioned datapoints and augmented datapoints. While each datapoint is unique, in the augmented dataset some datapoints may perform similar functions with parameter variations in the script. For instance, the augmented has few instances of sizing a gate and the gate sized can be different between datapoints.  Although these datapoints are distinct, they are categorized as ``augmented" due to their similar functionalities but differing script parameters. Two methods are applied to enrich the existing data and categorize them as ``Augmented'' in Table~\ref{tbl:stats}: 

\begin{enumerate}
    \item \emph{\textbf{Paraphrasing prompts}}: The work by Chang et al.~\cite{chang2023chipgpt} indicates that different prompts will affect the quality of the generated code by LLMs. Therefore, we have paraphrased prompts to link the same script to multiple semantic meanings to mitigate such effects.
    \item \emph{\textbf{Variable and parameter changes}}: The balance of datapoints across the different categories will highly affect the training process of data-driven algorithms. To keep a balanced dataset where each unique function has a sufficient representation, we augment the dataset by changing variable names in the script and input parameters.
\end{enumerate}

Some of the flow-based data are generated using automated scripting techniques. For instance, different contextual prompts are generated for each physical design stage, and domain experts generate different settings and implementations of each physical design stage. With permuting different stages in the flow, unique physical design flow scripts are created. We also use the same augmentation techniques to enrich the flow-based dataset. The DB query and design modification datasets are created using a specific design as an example and are based on manually designed prompts and corresponding scripts. These prompts and scripts are created by six individuals, inherently creating different semantic representations of the prompts based on individual preference and encapsulating different scripting styles of each individual 

\subsection{Validating the Dataset}
\label{validating_data}
\noindent 
{\bf Question-answer dataset}
To ensure the quality of the collected data, data samples are only taken from responses provided by OpenROAD experts, validated as correct by OpenROAD experts, or sampled from official OpenROAD documentation. Further, prompts and responses were proofread and adjusted to correct grammar, although some prompts were not adjusted to represent realistic user prompts better.

\noindent
{\bf Prompt-script dataset}
Each datapoint in the prompt-script dataset undergoes validation by a domain expert, who executes the script within OpenROAD and evaluates the correctness of the output against the specified prompt. This validation process mirrors the methodology employed in unit testing. Given that the prompts are manually crafted and the scripts are tailored for each prompt within the OpenROAD application, every datapoint is inherently correct by construction.

\section{Impact of the Corpus}
To demonstrate the impact of the corpus, we run a simple experiment by comparing different versions of ChatGPT, shown in Table~\ref{table:appendix_experiment}. For each of the datasets, we separate the dataset into 95\% training and 5\% validation. Then, we use the training set to fine-tune ChatGPT3.5 using the ``auto'' fine-tuning settings. Lastly, we evaluate each model with the validation set to determine the performance on each task. We perform fine-tuning and validation separately for each dataset.

\begin{table}[t]
\centering
\caption{Comparison between ChatGPT3.5/4 and finetuned ChatGPT3.5}
\label{table:appendix_experiment}
\resizebox{\linewidth}{!}{%
\begin{tabular}{||l||r|r||r|r|r||} 
\hhline{|t:=:t:==:t:===:t|}
\multicolumn{1}{||c||}{\multirow{2}{*}{\textbf{Model}}} & \multicolumn{2}{c||}{\textbf{Prompt-script}} & \multicolumn{3}{c||}{\textbf{Question-answer}} \\ 
\cline{2-6}
\multicolumn{1}{||c||}{} & \multicolumn{1}{c|}{\textbf{True}} & \multicolumn{1}{c||}{\textbf{False}} & \multicolumn{1}{c|}{\textbf{True}} & \multicolumn{1}{c|}{\textbf{Partial}} & \multicolumn{1}{c||}{\textbf{False}} \\ 
\hhline{|:=::==:b:===:|}
\textbf{ChatGPT3.5} & 0 & \multicolumn{1}{r||}{50} & 4 & 17 & 9 \\ 
\hline
\textbf{ChatGPT4} & 3 & 47 & 6 & 17 & 7 \\ 
\hhline{|:=::==::===:|}
\begin{tabular}[c]{@{}l@{}}\textbf{ChatGPT3.5 fine-tuned}\\\textbf{ with prompt-script~}\end{tabular} & 22 & 28 & -- & -- & -- \\ 
\hline
\begin{tabular}[c]{@{}l@{}}\textbf{ChatGPT3.5 fine-tuned}\\\textbf{ with question-answer}\end{tabular} & -- & -- & 26 & 3 & 1 \\
\hhline{|b:=:b:==:b:===:b|}
\end{tabular}
}
\vspace{-5mm}
\end{table}

\begin{figure}
    \centering
    \includegraphics[width= 0.92\linewidth]{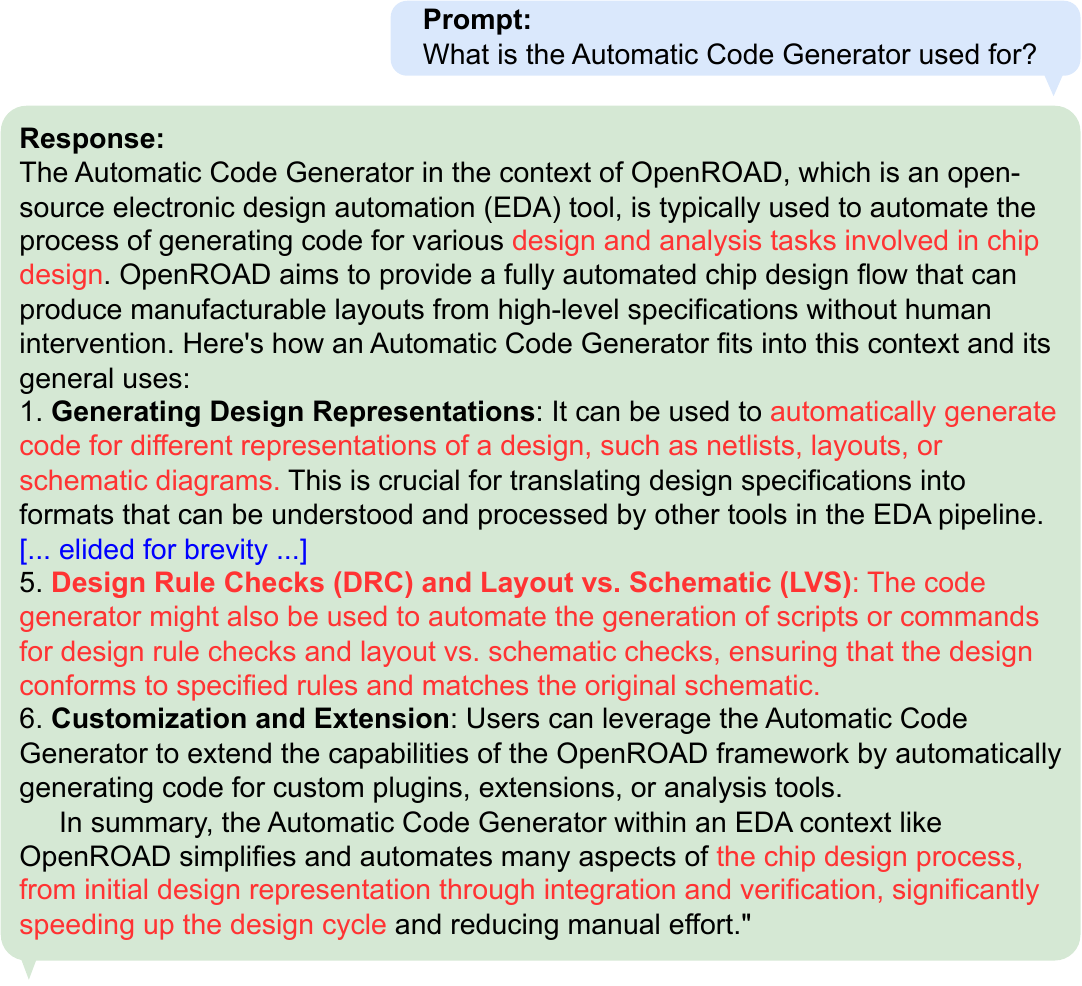}
    \vspace{-2mm}
    \caption{ChatGPT4 for question answering. Red text is factually incorrect.}
    \vspace{-4mm}
    \label{fig:qa_GPT4_example}
\end{figure}

\begin{figure}
    \centering
    \includegraphics[width= 0.92\linewidth]{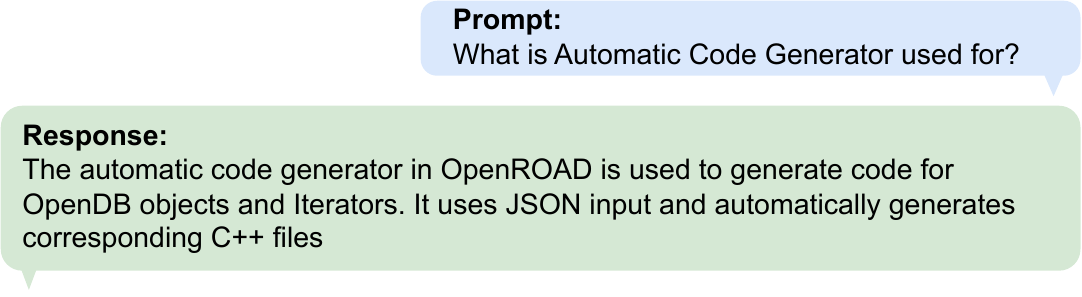}
    \vspace{-4mm}
    \caption{Example using fine-tuned ChatGPT3.5 for question answering.}
    \vspace{-6mm}
    \label{fig:qa_finetuned_GPT3.5_example}
\end{figure}

\begin{figure}[t]
    \centering
    \includegraphics[width= 0.92\linewidth]{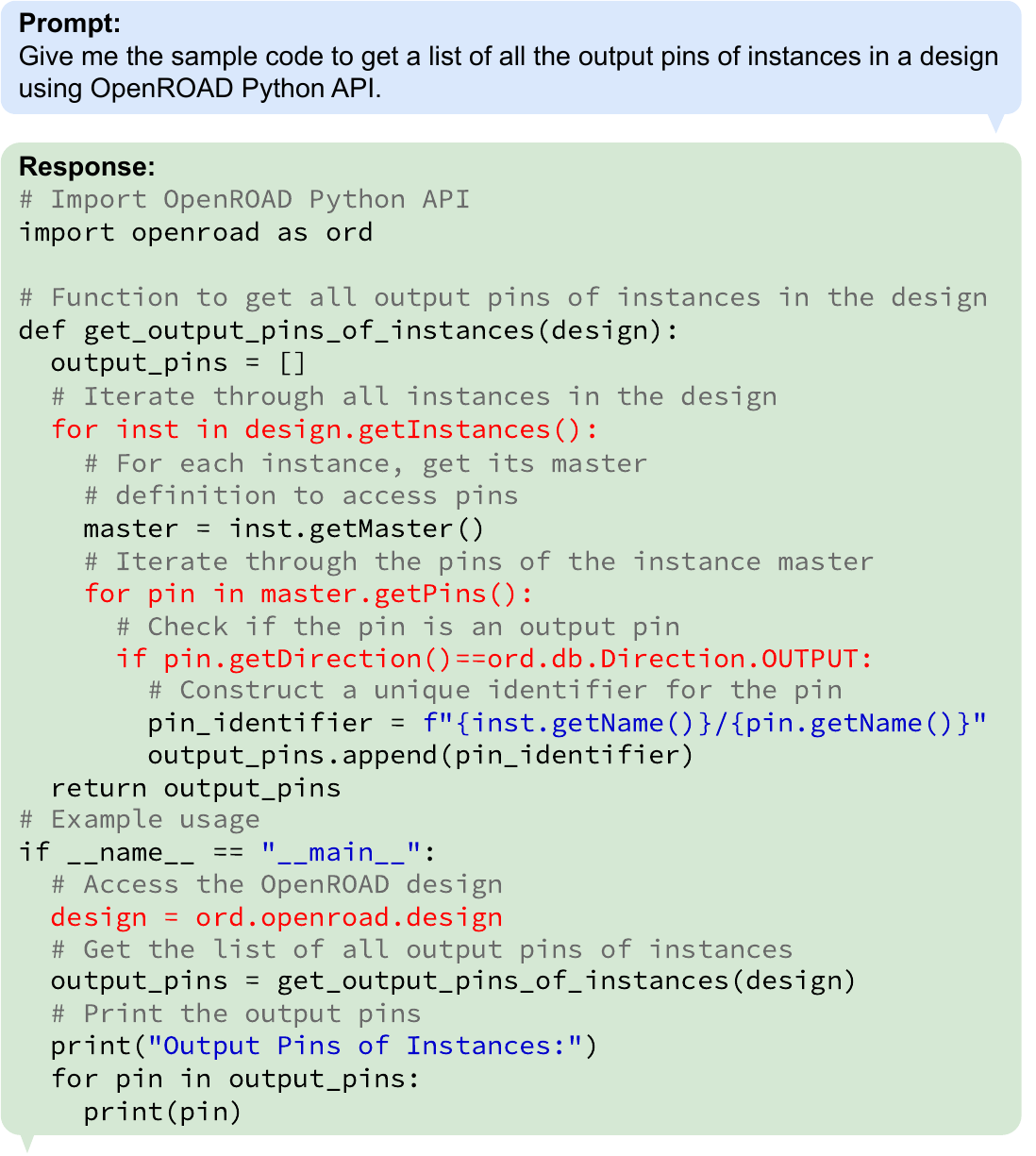}
    \vspace{-3mm}
    \caption{Example using ChatGPT4 for script generation. Red lines are incorrect.}
    \vspace{-4mm}
    \label{fig:ps_GPT4_example}
\end{figure}

\begin{figure}
    \centering
    \includegraphics[width= 0.92\linewidth]
    {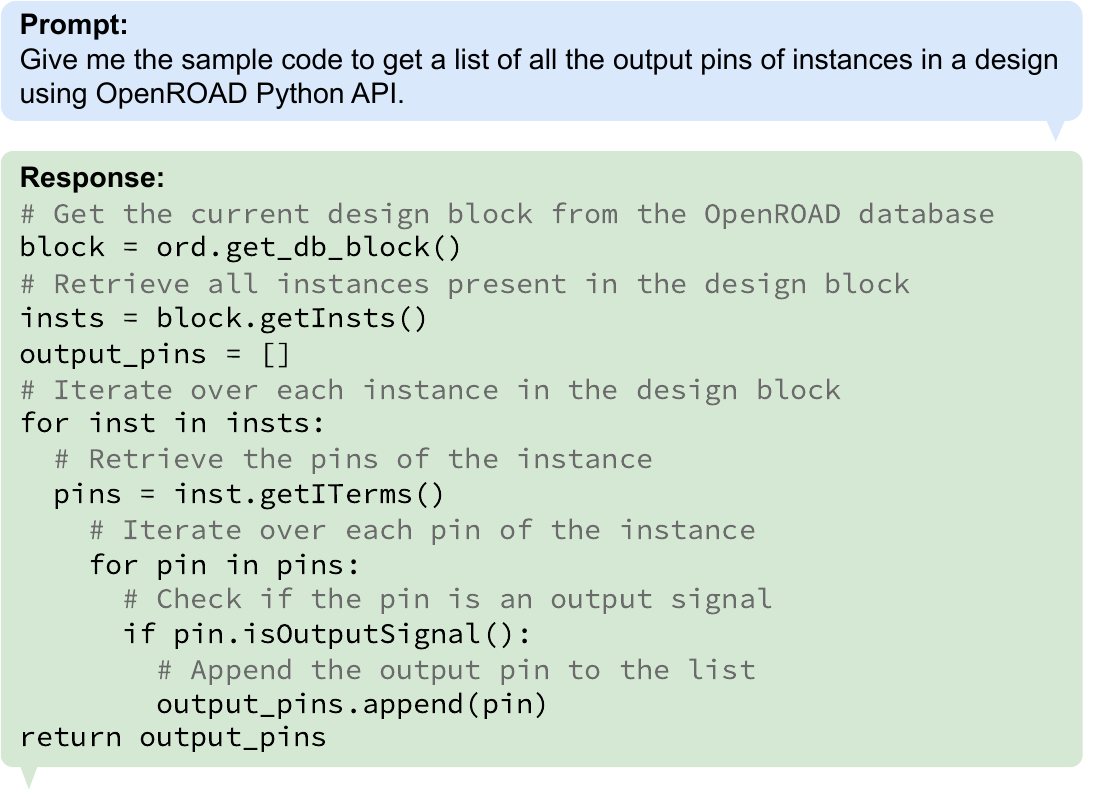}
    \vspace{-3mm}
    \caption{Example using fine-tuned ChatGPT3.5 for script generation.} 
    \vspace{-6mm}
    \label{fig:ps_finetuned_GPT3.5_example}
\end{figure}

\subsection{Question-Answer Dataset}
For the question-answer dataset, the responses are marked as true if the response contains only factual information and answers the question, partially true if the response answers the question correctly but also contains false information, and false if the response does not answer the question or provides only false information. These answers are judged by a consensus of OpenROAD experts. Fig.~\ref{fig:qa_GPT4_example} shows a False example where ChatGPT4 produces a response with factually incorrect information present and fails to answer the question correctly. Fig.~\ref{fig:qa_finetuned_GPT3.5_example} shows an example of a True response where the question is answered and contains factually correct information.

Table~\ref{table:appendix_experiment} demonstrates that on the question-answer task, ChatGPT3.5/4 does not do well. However, it is able to provide many partially correct answers by relying on general EDA knowledge from the foundation model. The fine-tuned ChatGPT3.5 performs much better by offering concise answers and answering the question directly. In this case, the experiment demonstrates that while foundation models can provide general answers, fine-tuning with the dataset can improve EDA-specific answers.

\subsection{Prompt-Script Dataset}
For the prompt-script dataset, the responses are marked as true only if the model generates a script that returns the correct result when executed in OpenROAD. All other responses are marked as false. 
Fig.~\ref{fig:ps_GPT4_example} shows a False example where ChatGPT4 produces code with incorrect syntax or hallucinated APIs (red lines). Fig.~\ref{fig:ps_finetuned_GPT3.5_example} shows a True example from a fine-tuned ChatGPT3.5 model where the generated script produces the correct output when executed in OpenROAD.

On the prompt-script task, ChatGPT3.5/4 do exceedingly poorly on the task with almost no correct responses. After fine-tuning, however, the accuracy increases to nearly half as shown in Table~\ref{table:appendix_experiment}. This result demonstrates that the prompt-script dataset can greatly improve performance on EDA script generation tasks. However, further development and research are warranted to improve accuracy.

%Such an application can make the integration of the OpenROAD Python API into existing workflows or future research much smoother as people can quickly look up sample codes, which is anticipated to reduce development time in engineering and unleash ML-EDA research.

%\begin{figure}[t]
%    \centering
%    \includegraphics[width= 0.92\linewidth]{figs/dataset_example4.drawio.pdf}
%    \caption{Example using native ChatGPT3.5.}
%    \label{fig:GPT3.5_example}
%\end{figure}

\section{Conclusion}
\label{sec:conclusion}
\noindent
This paper introduces EDA Corpus, a pioneering open-source dataset designed to facilitate the integration of LLMs into physical design through the OpenROAD framework. By offering over a thousand data points across question-answer and prompt-script formats, EDA Corpus marks a significant step towards filling the gap in domain-specific data for LLM training, promoting research in LLM-assisted physical design. Our evaluation shows that fine-tuning LLMs with EDA Corpus leads to improved performance on physical design-specific tasks (script generation or question and answer), highlighting the critical role of tailored datasets.

\bibliographystyle{misc/ieeetr2}
\bibliography{misc/bibfile}

%\clearpage
%\input{sec/7-appendix}
%\clearpage
\end{document}